\documentclass[journal,twoside,web]{ieeecolor}
\usepackage{tmi}
\usepackage{cite}
\usepackage{amsmath,amssymb,amsfonts}
\usepackage{algorithmic}
\usepackage{graphicx}
\usepackage{textcomp}
\usepackage{multirow}
\usepackage{subfigure}
\usepackage{makecell}
\usepackage{url}
\usepackage[colorlinks,linkcolor=red]{hyperref}
\usepackage[ruled,vlined]{algorithm2e}
\definecolor{darkgreen}{rgb}{0.3,0.7,0.3}
\definecolor{darkred}{rgb}{0.9,0.3,0.3}

\def\BibTeX{{\rm B\kern-.05em{\sc i\kern-.025em b}\kern-.08em
    T\kern-.1667em\lower.7ex\hbox{E}\kern-.125emX}}
\begin{document}

\title{Multimodal Medical Image Classification via Synergistic Learning Pre-training}

\author{Qinghua Lin, \emph{Student Member}, \emph{IEEE}, Guang-Hai Liu, Zuoyong Li, \emph{Senior Member}, \emph{IEEE}, \\ Yang Li, \emph{Senior Member}, \emph{IEEE}, Yuting Jiang and Xiang Wu
% and David Zhang, \emph{Life Fellow}, \emph{IEEE}
\thanks{This work was supported in part by National Natural Science Foundation of China (62471207), Natural Science Foundation of Fujian Province (2024J02029), Joint Funds for The Innovation of Science and Technology in Fujian province (2024Y9028, 2023Y9280), Open Project of Fujian Key Laboratory of Medical Big Data Engineering (KLKF202301). (Corresponding authors: Zuoyong Li; Xiang Wu.)}
\thanks{Qinghua Lin is with the College of Biomedical Engineering, Fudan University, Shanghai 200433, China (e-mail: akametris@163.com).}
\thanks{Guang-Hai Liu is with the College of Computer Science and Engineering, Guangxi Normal University, Guilin 541004, China (e-mail:
liuguanghai009@163.com). Zuoyong Li is with the Fujian Provincial Key Laboratory of Information Processing and Intelligent Control, School of Computer and Big Data, Minjiang University, Fuzhou 350121, China (e-mail: fzulzytdq@126.com).}
\thanks{Yang Li is with the Department of Automation Science and Electrical Engineering, Beihang University, Beijing 100191, China (e-mail: liyang@buaa.edu.cn).}
\thanks{Yuting Jiang is with the Department of Digestive Endoscopy, Fuzhou University Affiliated Provincial Hospital, Provincial Clinical Medical College of Fujian Medical University, 350001, Fuzhou, China (email:yutingjiang@fjmu.edu.cn). Xiang Wu is with the Department of Urology, Fuzhou University Affiliated Provincial Hospital, Provincial Clinical Medical College of Fujian Medical University, 350001, Fuzhou, China (email: tianyang0909@pku.org.cn).}
\thanks{This work has been submitted to the IEEE for possible publication. Copyright may be transferred without notice, after which this version may no longer be accessible.}
}

\maketitle

\begin{abstract}
Multimodal pathological images are usually in clinical diagnosis, but computer vision-based multimodal image-assisted diagnosis faces challenges with modality fusion, especially in the absence of expert-annotated data. To achieve the modality fusion in multimodal images with label scarcity, we propose a novel ``pretraining + fine-tuning" framework for multimodal semi-supervised medical image classification. Specifically, we propose a synergistic learning pretraining framework of consistency, reconstructive, and aligned learning. By treating one modality as an augmented sample of another modality, we implement a self-supervised learning pre-train, enhancing the baseline model's feature representation capability. Then, we design a fine-tuning method for multimodal fusion. During the fine-tuning stage, we set different encoders to extract features from the original modalities and provide a multimodal fusion encoder for fusion modality. In addition, we propose a distribution shift method for multimodal fusion features, which alleviates the prediction uncertainty and overfitting risks caused by the lack of labeled samples. We conduct extensive experiments on the publicly available gastroscopy image datasets Kvasir and Kvasirv2. Quantitative and qualitative results demonstrate that the proposed method outperforms the current state-of-the-art classification methods.
The code will be released at: \url{https://github.com/LQH89757/MICS}.

\end{abstract}

\begin{IEEEkeywords}
Multimodal, Medical image classification, Semi-supervised learning, Distribution shift, Deep learning.
\end{IEEEkeywords}

\section{Introduction}
\label{sec:introduction}

\begin{figure}[!t]
  \centering
  \includegraphics[width=\linewidth]{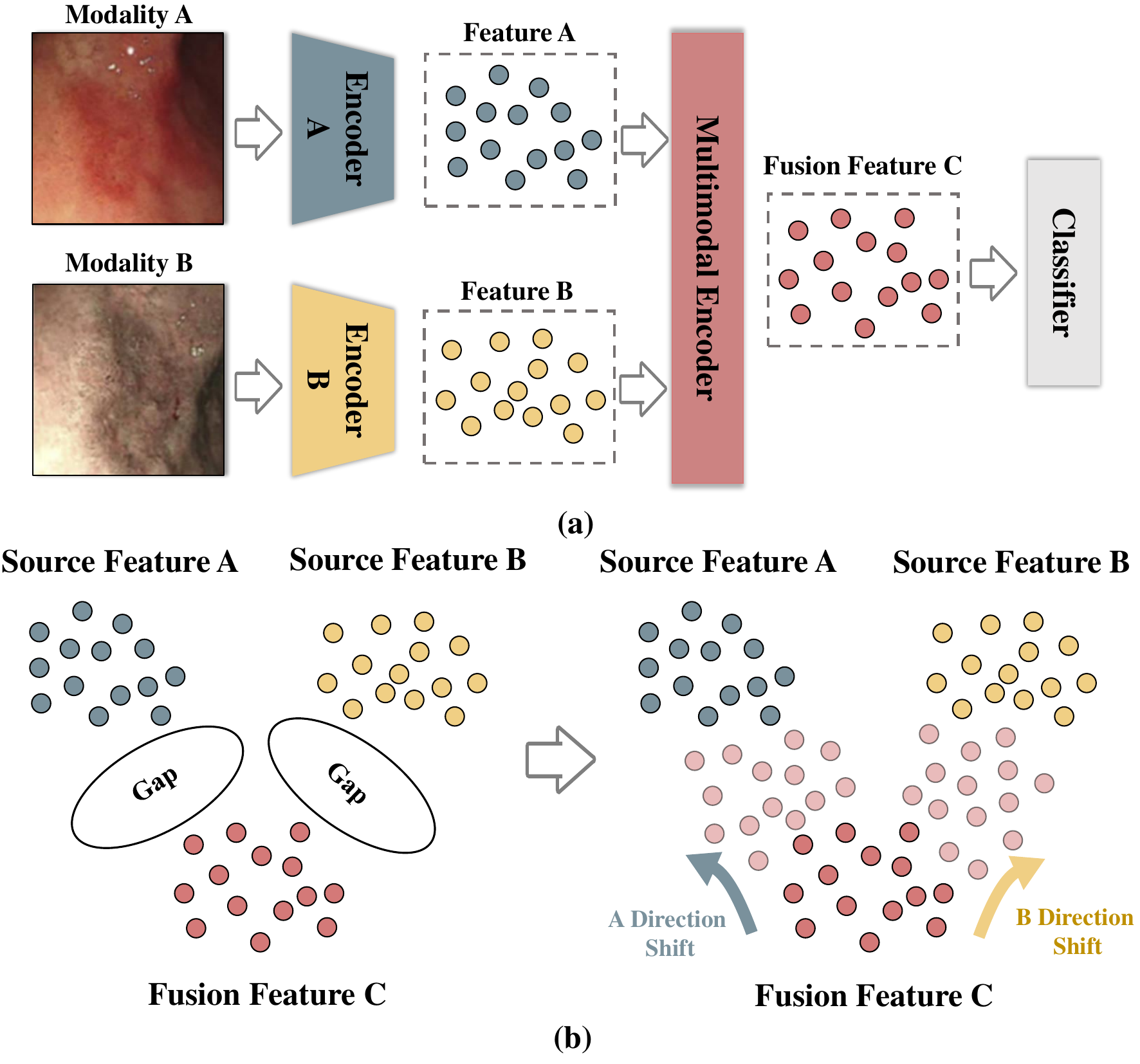}
  \caption{The proposed distribution shift method. (a)  Multimodal features fusion for classification during the fine-tuning stage. (b) The generation of shift directions based on feature distribution guides the augmentation of fused features.}
\label{fig:Intro}
\end{figure}

\begin{figure*}[!t]
  \centering
  \includegraphics[width=\linewidth]{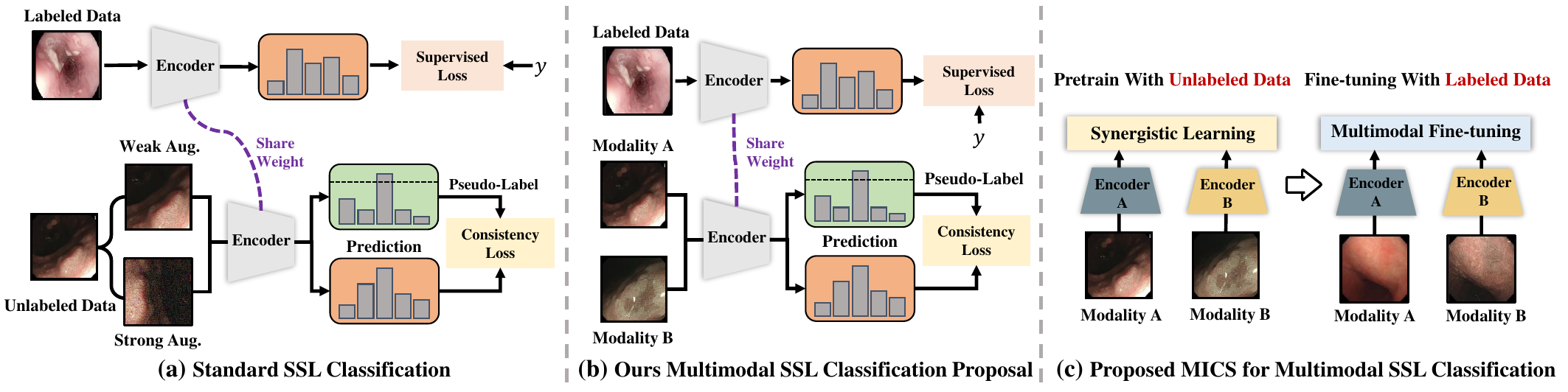}
  \caption{Comparison of Semi-Supervised Learning (SSL).}
\label{fig:Intro2}
\end{figure*}

\IEEEPARstart{W}{ith} the rapid development of medical imaging technology, medical images play an increasingly important role in clinical diagnosis and treatment \cite{XSF24,liyang1,liyang2,liyang3}. However, single-modal medical image information is limited and often fails to fully reflect the characteristics of lesions, which in turn affects the accuracy of diagnosis. By integrating different imaging techniques, multimodal medical images provide more diverse lesion information and offer a more reliable basis for clinical diagnosis and disease analysis \cite{TWM22}. For example, physicians usually use white light to inspect the patient's gastrointestinal tract and roughly identify suspicious areas. White light imaging uses the entire visible spectrum to produce red, green, and blue images, making it more sensitive in examining morphology, mucosa, and blood vessel distribution. This helps doctors detect gastric ulcers and tumors and is the current standard mode of endoscopic imaging. When a suspicious area is identified, the doctor switches to narrow-band imaging to further assess the possible type of condition. Narrow-band imaging uses specific filters to produce light in the blue and green wavelengths, enhancing the mucosal features and vascular structure of suspicious areas, which aids in the early detection of gastric cancer \cite{YSL17}.

However, due to the differences and redundancies between modalities, effectively fusing multimodal image information has become a pressing challenge \cite{LDM15}. Deep learning has made significant progress in medical image analysis in recent years, particularly demonstrating superior feature extraction and pattern recognition capabilities. As data-driven methods, deep models require considerable labeled data for training. The acquisition of data labeled by professional physicians is challenging and inefficient. Especially in multimodal tasks, doctors need to repeatedly switch between different modalities to ensure the accuracy of data annotation. Therefore, multimodal medical image tasks face the following two challenges:

(1) Multimodal medical images provide diverse lesion information, but effectively fusing features from different modalities remains challenging.

(2) Data annotation is more time-consuming and labor-intensive in multimodal image tasks than single-modal data.

To address these limitations, we rethink semi-supervised image classification methods and propose a \underline{\textbf{M}}ultimodal Medical \underline{\textbf{I}}mage \underline{\textbf{C}}lassification via \underline{\textbf{S}}ynergistic Learning Pre-training (MICS). As shown in \autoref{fig:Intro2}, current semi-supervised classification methods rely on consistency predictions between strong and weak augmentations of unlabeled images. Strong augmentation techniques \cite{Randaugment,ReMixMatch} have successfully perturbed natural images for consistency regularization (e.g., cats, airplanes). However, these standard augmentation techniques can easily destroy lesion features of medical images, and it is hard to find an appropriate augmentation combination \cite{ReCLR}. Inspired by these semi-supervised classification methods, we treat one modality of the same disease type as the optimal augmented sample for another modality for consistency prediction. Moreover, the scale of multimodal datasets is usually tiny, and the mutual influence between the training of labeled and unlabeled samples limits the model's optimization \cite{simmatch}. In view of these issues, we separate the training of unlabeled and labeled samples and propose a self-supervised pretraining method based on synergistic learning with consistency, reconstructive, and aligned learning. To achieve multimodal feature fusion on labeled samples, we propose a fusion fine-tuning method based on feature distribution shift.

Specifically, we use self-supervised learning for the pre-train stage. Consistency learning treats different modalities as distinct augmented samples and predicts their consistency to learn feature representations of unlabeled samples. Reconstructive learning, after randomly masking the original image, extracts features from different encoders to a unified decoder for image reconstruction, focusing on the local detail representations of original modalities. Aligned learning forces the two modalities to approach each other in the high-dimensional feature space through instance-level contrastive learning, further enhancing the similarity of paired image representations.

During the fine-tuning stage, we follow a multimodal fusion framework to fuse feature representations from different modalities. To alleviate overfitting risks caused by a small number of labeled samples, we propose an implicit augmentation method based on a shift vector dictionary (SVD). We use pre-trained weights to construct an SVD for modality clustering. SVD treats the output of the feature by the multimodal encoder as prototypes and obtains shift vectors. Then, according to shift vectors, the fusion features generate perturbed features with a feature distribution close to the initial modality. As shown in \autoref{fig:Intro}, the augmented samples fill the gap between the initial and fused modalities in the latent space, compensating for the model's perceptual differences between the fusion and original modalities.

Our contributions are summarized as follows:

\begin{itemize}
\item To the best of our knowledge, we propose the first multimodal-based semi-supervised medical image classification method by rethinking the implementation of multimodal in semi-supervised learning.

\item We propose a synergistic learning pretraining framework that combines consistency, reconstructive, and aligned learning, effectively enhancing the model's multimodal representation capability in a self-supervised way.

\item We propose a multimodal fusion fine-tuning method based on feature distribution shift, effectively alleviating the overfitting risk caused by the lack of labeled samples and mitigating the model's perceptual differences between the fusion and initial modalities.

\item We conduct extensive experiments using two public gastroscopy image datasets, where the paired modality images are generated from WtNGAN \cite{wtngan} to verify the feasibility of using multimodal images as augmented samples. The experimental results show that the proposed method achieves promising results in multimodal medical image classification.

\end{itemize}

\section{Related Work}
In this section, we first review several advanced semi-supervised classification methods, then introduce the current state of the multimodal domain, and finally focus on some works related to multimodal fusion.

\subsection{Semi-Supervised Classification}
Semi-supervised learning aims to achieve performance comparable to supervised learning by leveraging a small amount of labeled data and a large amount of unlabeled data. Mainstream semi-supervised classification methods primarily integrate pseudo-labeling and consistency regularization to learn feature representations from unlabeled samples. For example, FixMatch \cite{fixmatch} sets a fixed threshold to generate pseudo-labels from weakly augmented unlabeled samples and enforces consistency by predicting these pseudo-labels on their strongly augmented versions. FlexMatch \cite{flexmatch} introduces curriculum pseudo-labeling to flexibly adjust thresholds for different classes. FreeMatch \cite{freematch} dynamically adjusts the pseudo-labeling threshold based on the model's learning state and introduces class fairness regularization to reduce sample bias. SimMatch \cite{simmatch} combines contrastive self-supervised pretraining with consistency regularization, narrowing the gap between supervised and semi-supervised learning during fine-tuning. PEFAT \cite{pefat} enhances semi-supervised classification performance for medical images through adversarial training from the perspective of loss distribution.
%Pseudo-Label \cite{Pseudo-label} generates pseudo-labeled samples from high-confidence unlabeled samples. ReMixMatch \cite{ReMixMatch} encourages the marginal distribution of predictions for unlabeled samples to align with the true marginal distribution through distribution alignment and augmentation anchoring.

\begin{figure*}[!t]
  \centering
  \includegraphics[width=\linewidth]{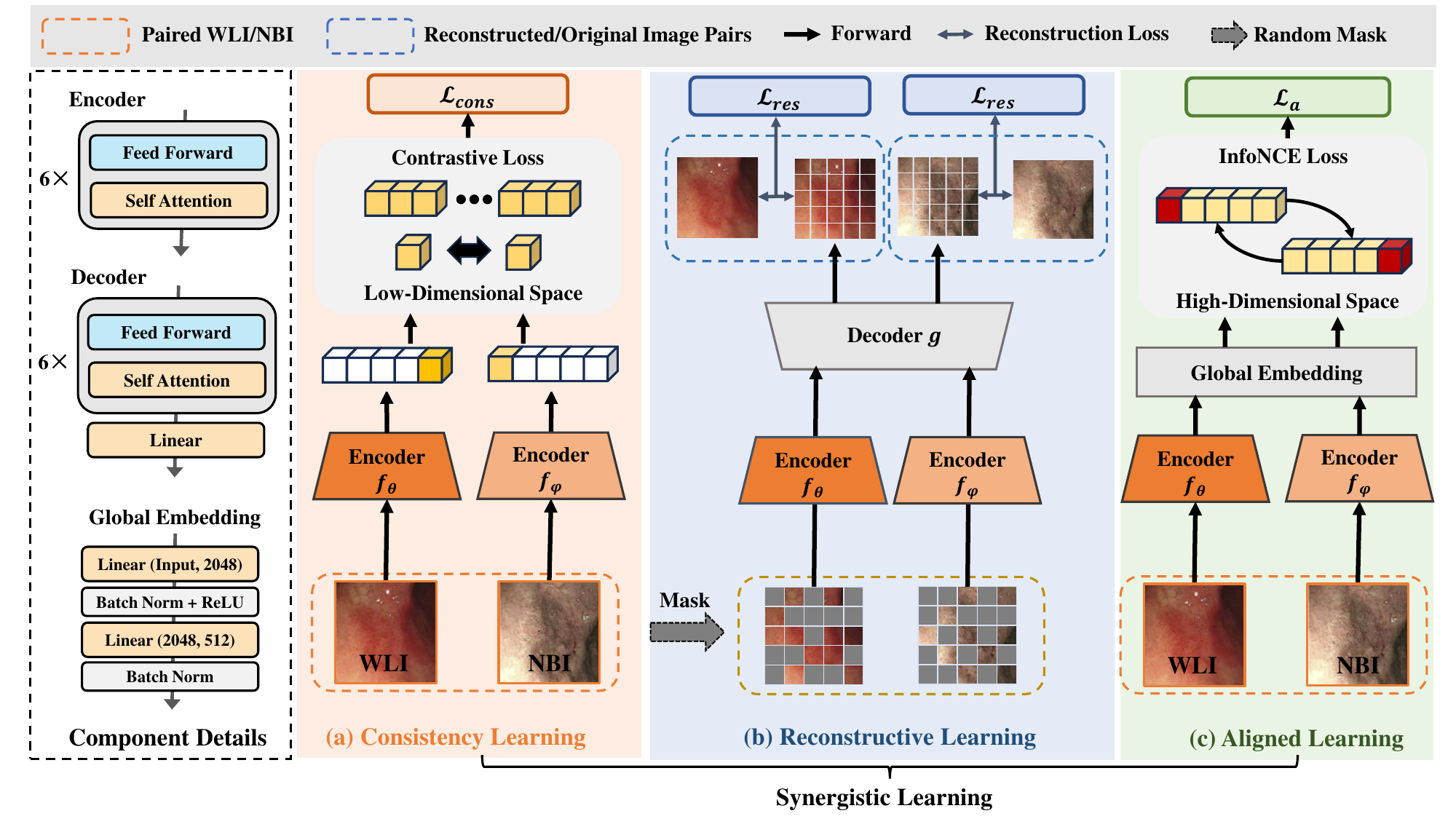}
  \caption{The overall framework of synergistic learning pre-train: (a)  Consistency learning constrains the latent representations of paired images in low-dimensional space to ensure Consistency. (b) Reconstructive learning randomly masks the original images and reconstructs them using a unified decoder based on the output features of single-modal encoders. (c) Aligned learning aims to align more complex and detailed single-modal representations in high-dimensional space. Synergistic learning provides robust single-modal representation capabilities for multimodal feature fusion models.}
\label{fig:Pretrain}
\end{figure*}

\subsection{Multimodal Task}
Multimodal visual tasks have garnered significant attention from researchers in recent years, as multimodal representations effectively leverage information from different modalities to improve downstream task performance. For example, ALBEF \cite{ALBEF} aligns feature representations of images and text through cross-modal attention.  DeepGuide \cite{deepguide} utilizes knowledge learned from a superior modality to guide the use of an inferior modality, effectively improving diagnostic performance for the latter. FusionM4Net \cite{fusionm4net} employs a two-stage multimodal learning approach to effectively fuse clinical and dermoscopic image representations at the feature level. GiMP \cite{GIMP} incorporates high-order correlation modeling, introducing a group multi-head self-attention gene encoder to capture the global structure of gene expression. FusAtNet \cite{fusatnet} utilizes multispectral and hyperspectral images for land cover classification. MSAN \cite{MSAN} focuses on modality-specific features, using two attention modules to guide eye disease classification based on fundus and OCT images.

% CCR-Net \cite{CCR-Net} performs cross-modal reconstruction of remote sensing data to learn more compact feature fusion representations.

\subsection{Multimodal Image Fusion}
Image fusion combines information from heterogeneous images to generate an image with rich details \cite{LRA20}. MATR \cite{TWM22} introduces a globally complementary context-adaptive modulation convolution to fuse multimodal medical images. ITFuse \cite{itfuse} employs interactive attention to merge complementary information from infrared and visible light images, effectively leveraging the shared attributes of heterogeneous images. TGFuse \cite{tgfuse} integrates features extracted by Transformers with shallow features from convolutional neural networks, refining cross-channel interactions within the spatial domain. 

% MDL-Net \cite{MDL-Net} enhances multimodal fusion representations from global, local, and latent learning perspectives, improving early diagnosis of Alzheimer’s disease.

Although semi-supervised classification has achieved performance close to supervised classification, it heavily relies on large-scale data and sufficient iterations. However, compared to natural images, acquiring medical imaging data is often more challenging. As a result, semi-supervised models face limitations in learning feature representations from unlabeled samples with small-scale medical imaging datasets. Considering clinicians typically analyze diseased areas using different imaging modalities and the advancements in multimodal tasks within computer vision, we propose MICS to enhance the benchmark for semi-supervised multimodal medical image classification.

\section{Method}
The proposed MICS consists of three stages: the pre-train stage, the dictionary construction stage, and the multimodal fusion fine-tuning stage. The pre-train stage includes consistency learning, reconstructive learning, and aligned learning. In the dictionary construction stage, a Shift Vector Dictionary (SVD) is built based on the feature representations learned from the pre-train stage. During the multimodal fusion fine-tuning stage, the model is initialized with pre-trained weights, while additional training samples are generated using the distribution shifts produced by the SVD.

\subsection{Synergistic Learning Pretrain}
The proposed synergistic learning pretrain framework is illustrated in \autoref{fig:Pretrain}. At the pretrain stage, we extract features from different modalities using single-modal encoders. For the input paired modality images $X=\{w,n\}$, the collaborative learning framework employs encoders $f_{\theta}$ and $f_{\varphi}$ to extract the white-light image features $z^{w}$  and narrow-band image features $z^{n}$, respectively. These latent features are first aligned in a low-dimensional feature space through consistency learning and calculate the consistency loss $\mathcal{L}_{cons}$. Then, we randomly mask $X=\{w,n\}$ as $X_m=\{w_m,w_n\}$. . The masked images are then processed by the encoders $f_{\theta}$ and $f_{\varphi}$ to extract features $z_{\mathrm{mask}}^w$ and $z_{\mathrm{mask}}^n$, which are subsequently reconstructed by a shared decoder $g$ and compute the reconstruction loss $\mathcal{L}_{res}$. Finally, to align the high-dimensional feature representations of different modalities, $z^{w}$ and $z^{n}$ are projected into a high-dimensional latent space using a global embedding $\mathrm{GloE}(\cdot)$ for an alignment loss $\mathcal{L}_{a}$.

\subsubsection{Consistency Learning} 
Inspired by image-text contrastive learning \cite{ALBEF}, we propose a consistency Learning method designed for use prior to the fusion of different modalities. Similar to MoCo \cite{MOCO}, the features of the most recently selected $K$ image pairs $T=\{t^w,t^n\}$ are stored in two separate queues as negative samples. During training, these $K$ image feature pairs are passed through a projection head $p$, which maps them into a low-dimensional space, and their similarity is computed as follows:

\begin{equation}
    \begin{gathered}
sim_{wn}^m=\frac{zm^w\times t^{n^{\prime}}}\tau,sim_{nw}^m=\frac{zm^n\times t^{w^{\prime}}}\tau, \\
sim_{wn}=\frac{z^w\times t^{n^{\prime}}}{\tau},sim_{nw}=\frac{z^n\times t^{w^{\prime}}}{\tau} ,
    \end{gathered}
\label{eq.1}
\end{equation}
where $zm^{w}$ and $zm^{n}$ represent the image features extracted by the momentum model, $t^{n^{\prime}}=cat((zm^n)^{\mathrm{T}},t^n)$, $t^{w^{\prime}}=cat((zm^w)^{\mathrm{T}},t^w)$, and $\tau$ denotes the temperature parameter. Subsequently, we construct a target similarity matrix $\mathbb{I}$ and compute the target similarity as follows:

\begin{equation}
    \begin{gathered}
        sim{\_}t_{wn}=\alpha\cdot softmax(sim_{wn}^m)+(1-\alpha)\cdot\mathbb{I},\\
        sim{\_}t_{nw}=\alpha\cdot softmax(sim_{nw}^m)+(1-\alpha)\cdot\mathbb{I}.
    \end{gathered}
\end{equation}
Therefore, the similarity between the features of the current image pair and the features in the momentum queue is defined as:
\begin{equation}
    \begin{gathered}
    \mathcal{L}_{wn}=-\frac1N\sum_i\sum_j\log\_\mathrm{softmax}(sim_{wn})\cdot sim\_t_{wn}, \\
    \mathcal{L}_{nw}=-\frac1N\sum_i\sum_j\log\_\mathrm{softmax}(sim_{nw})\cdot sim\_t_{nw}.
    \end{gathered}
\end{equation}
The final consistency loss $\mathcal{L}_{cons}$ is defined as:
\begin{equation}
\mathcal{L}_{dis}=\frac{\mathcal{L}_{wn}+\mathcal{L}_{nw}}2.
\end{equation}

\subsubsection{Reconstructive Learning}
The masked autoencoder \cite{MAE} is considered a practical self-supervised learning approach. Therefore, we introduce a generative learning strategy to capture local detail representations across different modalities in the synergistic learning pre-train stage. Different from \cite{MAE}, we employ a unified decoder $g$ for multimodal tasks to facilitate interaction between modal features, which means that regardless of whether the image features extracted by encoder $f_{\theta}$ and $f_{\varphi}$, the decoder can reconstruct the original images. The motivation of reconstructive learning is analogous to adversarial learning, where we aim for different modalities to align only in critical regions within the feature space rather than achieving complete consistency. The unified decoder $g$ helps retain modality-specific feature representation capabilities for each encoder.

Specifically,  we reconstruct the masked features $z_{\mathrm{mask}}^w$ and $z_{\mathrm{mask}}^n$ into the original images $w^{r}$ and $n^{r}$: 
\begin{equation}
z_{mask}^w=f_\theta(\text{Mask}(w,\sigma)),z_{mask}^n=f_\varphi(\text{Mask}(w,\sigma)),
\end{equation}
where $\text{Mask}(\cdot)$ denotes random masking, and $\sigma=0.75$ represents the masking ratio. Then, through the decoder $g$, the reconstruction loss $\mathcal{L}_{res}$ is defined as:
\begin{equation}
L_{res}=\|w,g(z_{mask}^w)\|_2+\|n,g(z_{mask}^n)\|_2.
\end{equation}

\begin{figure*}[!t]
  \centering
  \includegraphics[width=\linewidth]{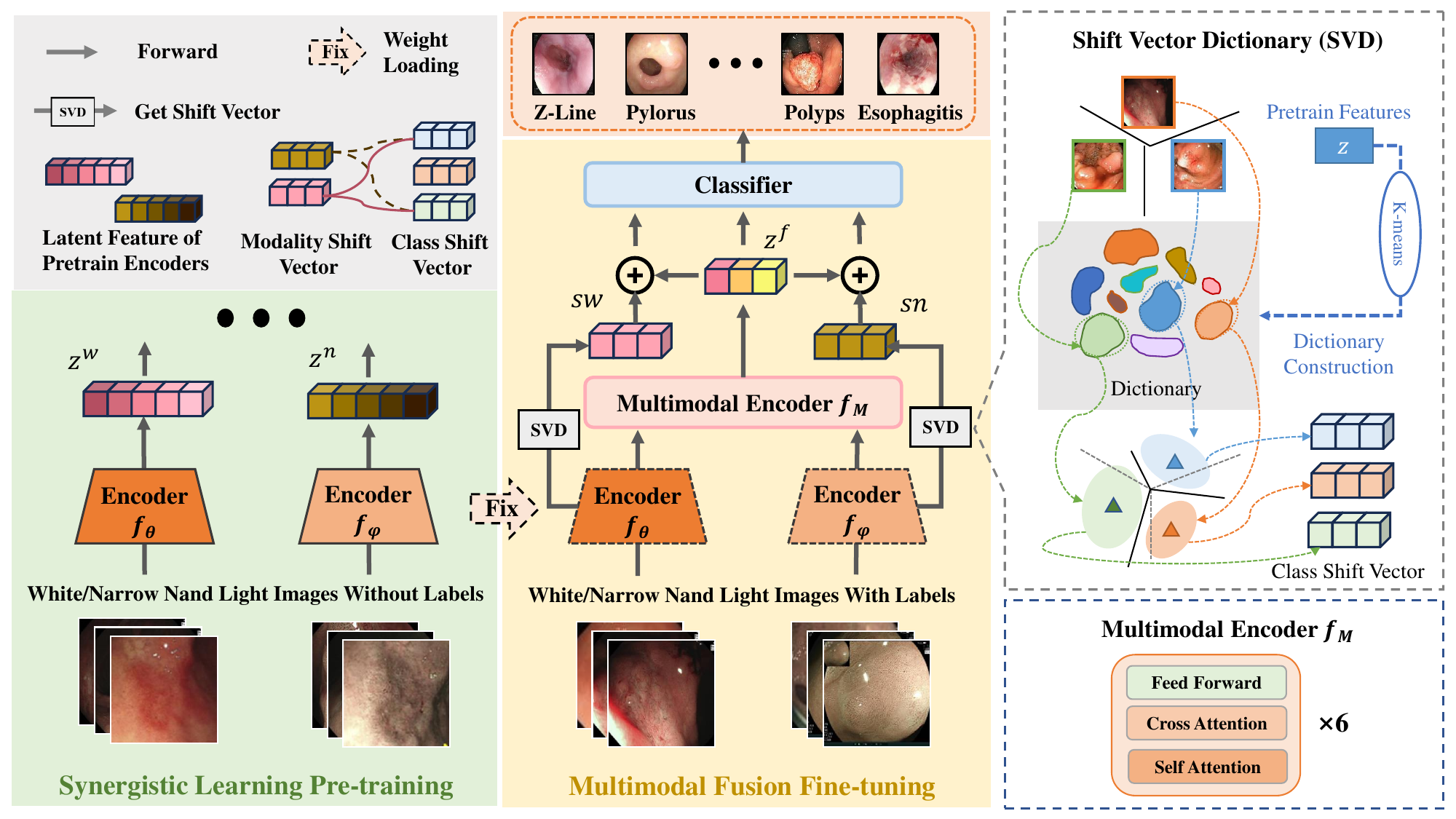}
  \caption{The overview of proposed multimodal fusion fine-tuning framework. At the fine-tuning stage, the encoders load the weights from pre-training. A small number of labeled multimodal samples are encoded by $f_{\theta}$ and $f_{\varphi}$, and then fed into the multimodal encoder $f_{M}$ to learn the fused representation. Subsequently, the multimodal features $z^{f}$ are perturbed based on the shift vectors generated from the shift vector dictionary.}
\label{fig:MMF}
\end{figure*}

\begin{figure*}[!t]
  \centering
  \includegraphics[width=\linewidth]{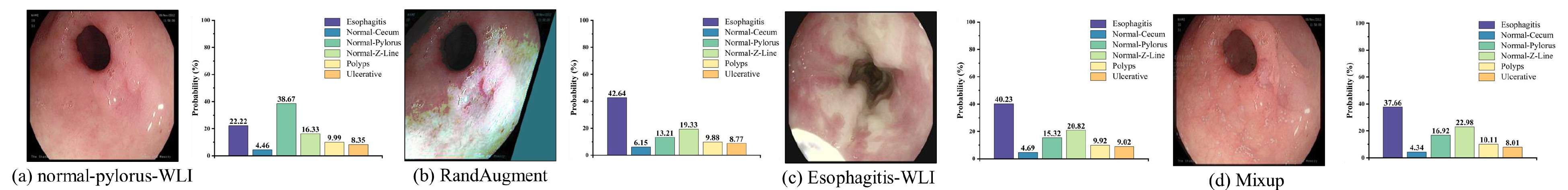}
  \caption{The sample of white light image. (a) Normal-pylorus sample. (b) The sample (a) after randaugment. (c) Esophagitis sample. (d) The sample (a) mixes with Polyps using a mixup.}
\label{fig:Aug}
\end{figure*}

\subsubsection{Aligned Learning}
The research \cite{HVCLIP} reveals the role of high-dimensional features in domain adaptation. For medical images, high-dimensional vectors facilitate the model to enrich feature representations and recognize subtle lesions. Specifically, inspired by the research \cite{LQA24}, we set a global embedding to project $z^{w}$ and $z^{n}$ into high-dimensional latent feature space vectors $Z^{w}$ and $Z^{n}$. Aligned learning optimizes the InfoNCE \cite{InfoNCE} loss as the objective through instance-level contrastive learning to align representations. Similar to \autoref{eq.1}, we compute the similarity between the embedding vectors of WLI and NBI as follows:
\begin{equation}
\begin{gathered}
S(w_i,n_j)=\frac{\exp{(cosine(Z_i^w,Z_j^n)/\tau)}}{\sum_{b=1}^B\exp{(cosine(Z_i^w,\tilde{Z}_b^n)/\tau)}}, \\
S\big(n_i,w_j\big)=\frac{\exp{(cosine(Z_j^n,Z_i^w)/\tau)}}{\sum_{b=1}^B\exp{(cosine(Z_i^n,\tilde{Z}_b^w)/\tau)}},
\end{gathered}
\end{equation}
where $cosine(\cdot)$ denotes cosine similarity, $\tau$ is the temperature as in \autoref{eq.1}, $(Z_i^w,Z_j^n)$ represents positive instance pairs, $\tilde{Z}_b^w$ and $\tilde{Z}_b^n$ represent negative instances. Thus, the optimization objective of aligned learning is defined as follows:
\begin{equation}
\mathcal{L}_{a}=\frac{1}{B}\sum_{i=1}^{B}(\mathcal{H}(y_{i},S(w_{i},n))+\mathcal{H}(y_{i},S(n_{i},w))) ,
\end{equation}
where $\mathcal{H}(\cdot)$ denotes the cross-entropy, and $y_i\in\mathbb{R}^B$ represents the one-hot labels of the instance-level samples.

\subsubsection{Loss Function in Pretrain Stage}
The overall optimization objective of MICS during the pre-training stage is summarized as follows:
\begin{equation}
L_{pre}=\alpha^{dis}L_{dis}+\alpha^{res}L_{res}+\alpha^{a}L_{a},
\end{equation}
where $\alpha^{dis}$, $\alpha^{res}$ and $\alpha^{a}$ represent the weighting coefficients for the corresponding losses.

\begin{figure*}[t]
  \centering
  \includegraphics[width=0.85\linewidth]{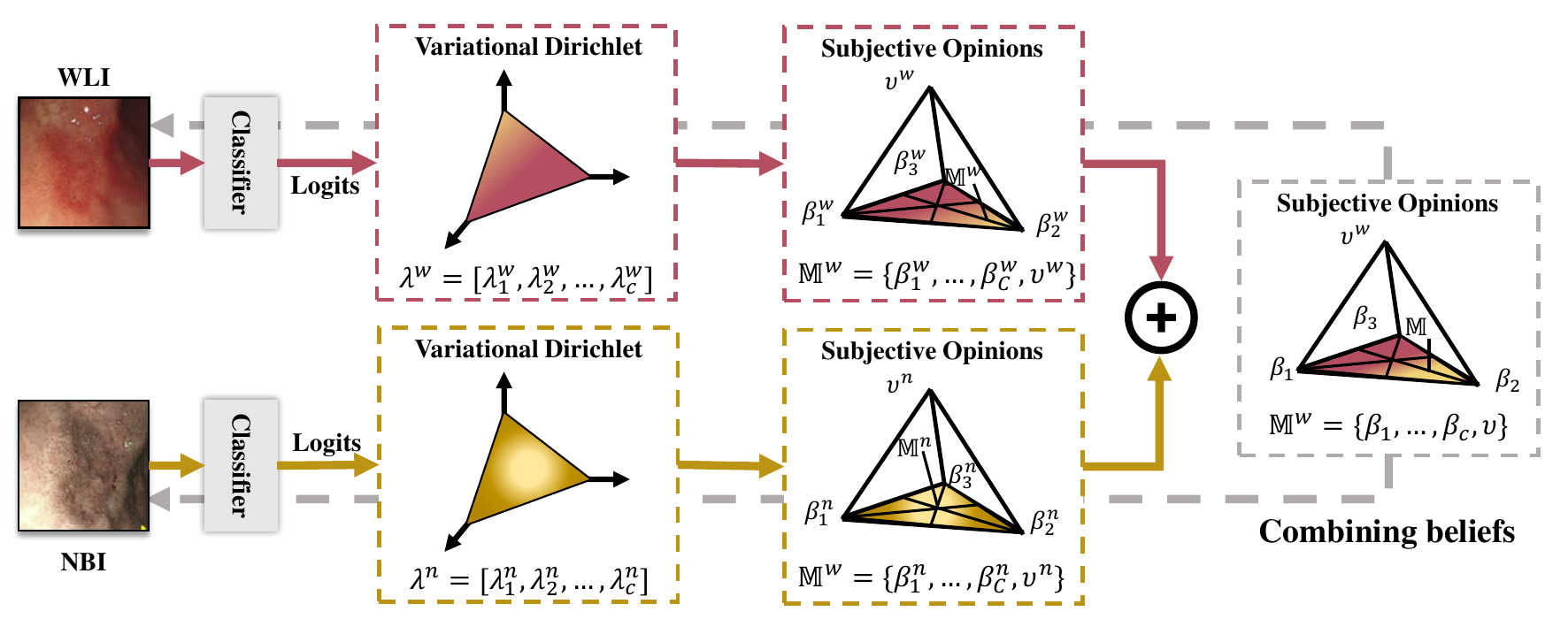}
  \caption{Overview of Dynamic Evidential Fusion \cite{ETMC}.}
\label{fig:ETMC}
\end{figure*}

\subsection{Shift Vector Dictionary}
Previous explicit augmentation methods (e.g.,  mixup \cite{mixup} and randaugment \cite{Randaugment}) are prone to generate unreliable augmented samples for medical images. As shown in \autoref{fig:Aug}, we provide an example of a standard white-light image. In semi-supervised classification methods, the prediction of weakly augmented (a) is typically used as a pseudo-label, forcing the model to predict (b) as the same class. The ambiguous consistency between (a) and (b) affects the model's prediction for (c). The augmented samples with randaugment intuitively resemble real Esophagitis more closely, even if a well-trained ViT-B/16 model misclassifies them. Besides, the augmented sample of mixup from Normal-Pylorus and Polyps is incorrectly identified as Esophagitis.

Therefore, we consider designing an implicit augmentation scheme that requires no prior knowledge. On the one hand, it avoids introducing potentially ambiguous semantic information. On the other hand, it transfers transferable knowledge from existing data. Inspired by \cite{JYT22}, we propose an implicit feature augmentation scheme based on the distribution shift of the base class dictionary. The difference between the proposed method and \cite{JYT22} is that the former is suitable for multimodal tasks, where the sample distribution of the base class dictionary generates shift components to guide the fusion features toward the initial modality, while the latter seeks the nearest set of samples within a single-modal dictionary’s clustering to join the training process.

Specifically, we fully leverage the pre-trained encoders $f_{\theta}$ and $f_{\varphi}$. Since the intermediate layers of the encoders operate independently during the pretraining stage, $f_{\theta}$ and $f_{\varphi}$ retain the ability to map the original modality distribution. First, the encoders extract features from unlabeled samples:
\begin{equation}
\begin{gathered}
z_w^*=f_\theta(w_q),q=\{1,2,...,Q\}, \\
z_n^*=f_\varphi(n_q),q=\{1,2,...,Q\},
\end{gathered}
\end{equation}
where $Q$ represents the total number of samples in the training set. Then, the extracted features are clustered into $C$ clusters by K-means:
\begin{equation}
\min\sum_{i=1}^Q\sum_{j=1}^C\mathbf{1}\{c_i=j\}\left\|z_i^*-\mu_j\right\|^2,
\end{equation}
where $c=\{c_1,c_2,...,c_Q\}$ represents the cluster labels and $c_i\in\{1,2,...,C\}$. $C$ is the number of classification categories, $\mu_{j}$ is the prototype vector for each cluster, and $1\{c_i=j\}$ is the indicator function. Then, for each cluster $j\in\{1,2,...,C\}$, the sample mean is calculated to update the prototype $\mu_{j}$ as:
\begin{equation}
\mu_j=\frac1{Q_j}\sum_{i:c_i=j}z_i^*,
\end{equation}
where $Q_j$ is the number of samples in cluster $j$. Therefore, the sample covariance matrix for cluster $j$ can be represented as:
\begin{equation}
\Sigma_{j}=\frac{1}{Q_{j}-1}\sum_{i:c_{i}=j}\bigl(z_{i}^{*}-\mu_{j}\bigr)\bigl(z_{i}^{*}-\mu_{j}\bigr)^{T}.
\end{equation}
Finally, we compute the shift vector dictionary based on the covariance matrix $\Sigma_{j}$. Sampling $P$ shift vectors $\{s_1^j,s_2^j,...,s_P^j,\}$ from a multivariate Gaussian distribution with mean $\mu_{j}$ and covariance $\Sigma_{j}$. The shift vector $s_p^j$ is defined as:
\begin{equation}
s_p^j{\thicksim}\mathcal{N}(\mu_j,\Sigma_{j}),p=1{,}2,...,P,
\end{equation}
The shift vector dictionary for all clusters is $SVD\in\mathbb{R}^{C\times P\times D}=\{\{s_p^j\}_{p=1}^P\}_{c=1}^C$, where $D$ represents the dimensionality of the original input features. 

\subsection{Multimodal Fusion Fine-tuning}
The multimodal fusion fine-tuning process is shown in \autoref{fig:MMF}. The fine-tuning stage includes the initial modality encoders $f_{\theta}$ and $f_{\varphi}$, provided by synergistic learning pretrain, and the multimodal encoder $f_{M}$. We fine-tune the model using a small amount of labeled data $L_X=\{l_w,l_n\}$. Specifically, $f_{\theta}$ and $f_{\varphi}$ extract the corresponding modality features $z^{lw}$ and $z^{ln}$, features are fused by $f_{M}$ to obtain the multimodal fused feature $z^{f}$. Then, based on the initial modality Shift Vector Dictionary constructed in Section III.B, we randomly select the prototypes $\mu=\{\mu^w,\mu^n\}$ and their corresponding shift vectors $s_{w}$ and $s_{n}$. Finally, $z^{f}$ is element-wise added to the shift vectors $s_{w}$ and $s_{n}$ to obtain the knowledge transfer-based distribution-shifted features $z^{wf}$ and $z^{nf}$. Therefore, the loss function for the proposed method during the multimodal fusion fine-tuning stage is defined as follows:
\begin{equation}
\begin{aligned}
\mathcal{L}_{f}&=\mathcal{H}\left(y,cls(z^{f})\right)+\mathcal{H}\left(y,cls(z^{f}+sw)\right)\\&+\mathcal{H}\left(y,cls(z^{f}+sn)\right),
\end{aligned}
\label{eq.15}
\end{equation}
where $y$ represents the class label, and $cls$ denotes the classification head.

We aim to achieve information interaction between modalities and remove redundant features during the fine-tuning stage through the cross-attention mechanism of the multimodal fusion encoder. Additionally, the proposed distribution-shift-based implicit data augmentation method mitigates the impact of ambiguous semantic information on model training and avoids introducing additional parameter computations. It expands the training data and alleviates the overfitting risk caused by the lack of labeled samples.

\subsection{Uncertainty-Based Evidential Fusion}
Inspired by multi-view classification, uncertainty-based evidential fusion can enhance the reliability of multimodal classification in addition to feature fusion. As shown in \autoref{fig:ETMC}, we introduce a dynamic evidential fusion method for multimodal medical image classification called Trusted Multi-View Classification (TMC) \cite{ETMC}. We did not modify the theory of TMC but extended it to the gastroscopy multimodal classification domain, providing new insights for subsequent multimodal classification tasks.

Specifically, we obtain the $C$ concentration parameters of the Dirichlet distribution for each modality sample in the fine-tuned model through TMC:
\begin{equation}
\lambda=[\lambda_1,\lambda_2,...,\lambda_c].
\end{equation}
Therefore, Dirichlet distribution is defined as $Dir(\sigma|\lambda)$, $\lambda$ is the mean of the Dirichlet distribution, Dirichlet strength can be defined as $DS=\sum_{c=1}^C\lambda_c$. Then, belief mass $\beta$ and uncertainty $\nu$ are defined as:
\begin{equation}
\begin{aligned}
\beta_c&=\frac{\lambda_c-1}{DS} ,\\
\nu&=\frac C{DS} ,
\end{aligned}
\end{equation}
where $C$ is the number of classification categories. For samples from two modalities, TMC generates $\beta_c^w$ and $\nu_c^w$ for WLI, and $\beta_c^n$ and $\nu_c^n$ for NBI. Therefore, the probability mass assignments for different modalities are defined as:
\begin{equation}
\begin{aligned}
\mathbb{M}^{w}=\{\{\beta_{c}^{w}\}_{c=1}^{C},\upsilon^{w}\}, \\
\mathbb{M}^{n}=\{\{\beta_{c}^{n}\}_{c=1}^{C},\upsilon^{n}\}.
\end{aligned}
\end{equation}
Given the quality of the fine-tuned model for each modality, these beliefs and uncertainties from different modalities are fused as:
\begin{equation}
\mathbb{M}=\mathbb{M}^w\oplus\mathbb{M}^n.
\end{equation}

Although $\mathbb{M}$ provides a decision-level fusion for the fine-tuned model, the external modality complementary optimization objectives at the representation layer can promote information interaction between modalities. Therefore, the final fine-tuning loss is defined as follows:
\begin{equation}
\mathcal{L}=\mathcal{L}_f+\mathcal{L}_{wn}+\mathcal{L}_{fuse},
\end{equation}
where $\mathcal{L}_f$ is shown in \autoref{eq.15}. $\mathcal{L}_{wn}$ is the fusion loss between white light samples and narrow-band light samples, defined as follows:
\begin{equation}
\begin{gathered}
\mathcal{L}_{wn}=\log p^w(y|\sigma^w)+\log p^n(y|\sigma^n) \\
-\theta\cdot KL[Dir(\sigma^w,\lambda^w)\|Dir(\sigma^w,[1,...,1])] \\
-\theta\cdot KL[Dir(\sigma^n,\lambda^n)||Dir(\sigma^n,[1,...,1])],
\end{gathered}
\end{equation}
where $p^w$ and $p^n$ represent the predicted probabilities for white light and narrow-band light images, respectively. $\sigma$ is the mean value of the Dirichlet Distribution for the corresponding modality, and $KL$ denotes the Kullback-Leibler Divergence. Additionally, the fusion sample loss $\mathcal{L}_{fuse}$ is defined as:
\begin{equation}
\begin{aligned}
 & \mathcal{L}_{fuse}=\log p(y|\sigma) \\
 & -\theta\cdot KL[Dir(\sigma,\lambda)||Dir(\sigma,[1,...,1])].
\end{aligned}
\end{equation}

\section{Experiment}

\begin{figure}[!t]
  \centering
  \includegraphics[width=\linewidth]{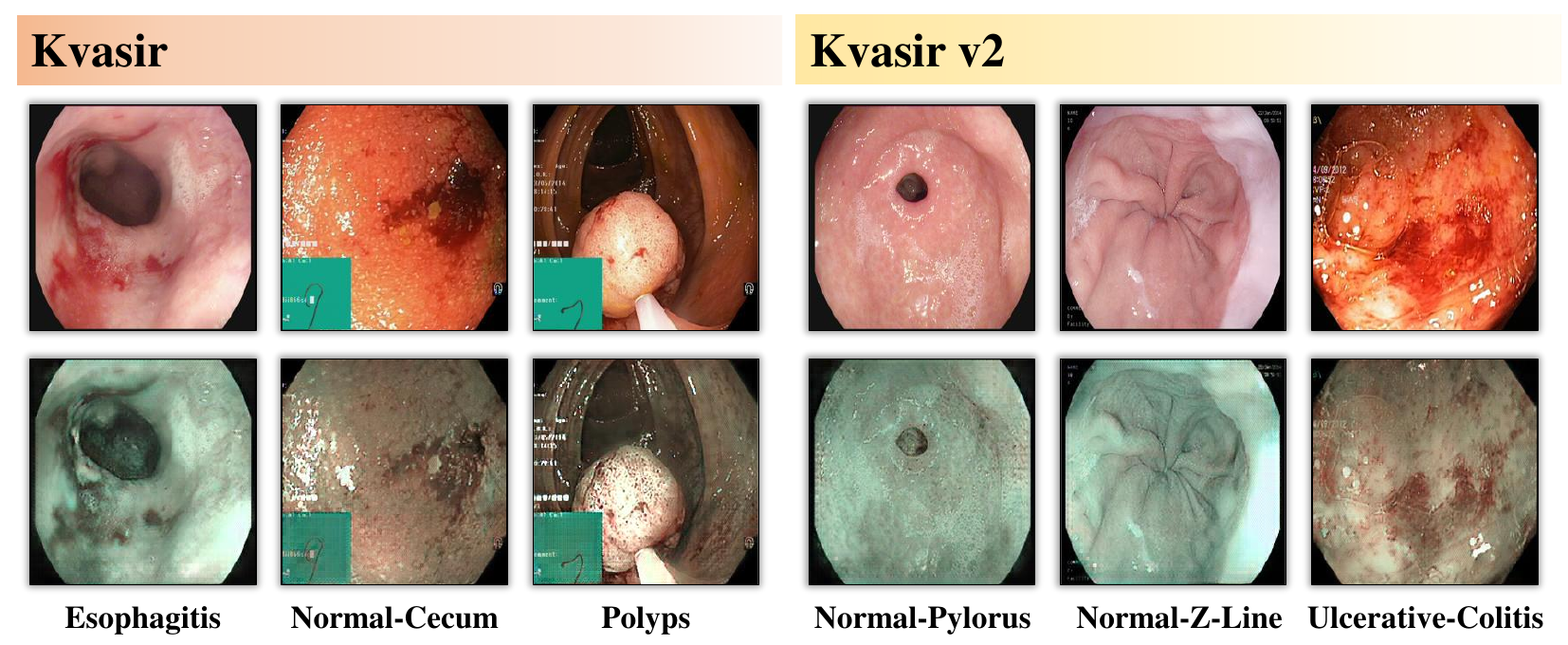}
  \caption{Sample examples from the Kvasir and Kvasirv2 datasets. The top row shows the real white light images, and the bottom row shows the paired narrow-band light images generated by WtNGAN.}
\label{fig:dataset}
\end{figure}

\subsection{Implementation Details}
At the pre-training stage, the initial learning rate is set to $1\times e^{-4}$ and gradually decays to $1\times e^{-5}$ with a cosine schedule. We use the AdamW \cite{AdamW} optimizer with a weight decay coefficient of 0.02 and a momentum parameter of 0.995, training for 100 epochs. At the fine-tuning stage, the parameter settings remain similar to the pre-training, except that the minimum learning rate $1\times e^{-6}$ and the decay rate 0.999. We implemented all methods in Python 3.9.13 and PyTorch 1.12.1. A computer equipped with an Intel Xeon Silver 4208 CPU and seven NVIDIA GeForce RTX 3090 GPUs for train and testing.

\begin{table}[!h]
\centering
\caption{The Detailed Composition Of Kvasir And Kvasirv2.}
\renewcommand{\arraystretch}{1.2}
\resizebox{0.4\textwidth}{!}{
\begin{tabular}{c|c|c|c}
\hline
                          & Categories  & Kvasir & Kvasirv2 \\ \hline
\multirow{3}{*}{Normal}   & Z-line      & 500    & 1,000    \\
                          & Pylorus     & 500    & 1,000    \\
                          & Cecum       & 500    & 1,000    \\ \hline
\multirow{3}{*}{Abnormal} & Esophagitis & 500    & 1,000    \\
                          & Polyps      & 500    & 1,000    \\
                          & Ulcerative-Colitis     & 500    & 1,000    \\ \hline
                          & Total       & 3,000  & 6,000    \\ \hline
\end{tabular}}
\label{tab:dataset}
\end{table}

\begin{table*}[!t]
\centering
\caption{Results of comparative experiments. Semi-supervised methods are trained in a standard way and a proposed multi-modality way. \textcolor{darkgreen}{Green} indicates that multi-modal training is improved compared to single-modality, and \textcolor{darkred}{red} indicates a decrease. \textbf{Bold} indicates the best result.}
\renewcommand{\arraystretch}{1.4}
\resizebox{\textwidth}{!}{
\begin{tabular}{c|c|cc|cccccccc}
\hline
\multirow{2}{*}{Method} & \multirow{2}{*}{Year} & \multicolumn{2}{c|}{Modality} & \multicolumn{4}{c}{Kvasir}                                                                                                                                                        & \multicolumn{4}{c}{Kvasir v2}                                                                                                                              \\ \cline{3-12} 
                        &                       & WLI           & NBI           & 50\% labels                          & 30\% labels                           & 10\% labels                           & \multicolumn{1}{c|}{5\% labels}                            & 50\% labels                          & 30\% labels                          & 10\% labels                          & 5\% labels                            \\ \hline
FixMatch                & 2020                  & \checkmark    &               & 71.67                                & 51.67                                 & 46.67                                 & \multicolumn{1}{c|}{44.67}                                 & 73.83                                & 71.75                                & 60.42                                & 49.75                                 \\
FlexMatch               & 2021                  & \checkmark    &               & 73.00                                & 67.33                                 & 52.33                                 & \multicolumn{1}{c|}{54.67}                                 & 73.00                                & 74.67                                & 63.50                                & 56.41                                 \\
Dash                    & 2021                  & \checkmark    &               & 70.33                                & 55.33                                 & 44.50                                 & \multicolumn{1}{c|}{40.17}                                 & 75.17                                & 71.75                                & 62.42                                & 46.08                                 \\
%DebiasPL                & 2022                  & \checkmark    &               & 76.17                                & 73.17                                 & 63.83                                 & \multicolumn{1}{c|}{59.67}                                 & 78.67                                & 74.25                                & 71.50                                & 70.00                                 \\
FreeMatch               & 2023                  & \checkmark    &               & 73.83                                & 69.17                                 & 52.33                                 & \multicolumn{1}{c|}{45.67}                                 & 75.17                                & 73.42                                & 69.25                                & 60.83                                 \\
SoftMatch               & 2023                  & \checkmark    &               & 73.33                                & 69.83                                 & 62.67                                 & \multicolumn{1}{c|}{54.33}                                 & 73.00                                & 71.50                                & 66.33                                & 60.41                                 \\
HABIT                   & 2023                  & \checkmark    &               & 73.17                                & 69.83                                 & 61.00                                 & \multicolumn{1}{c|}{54.83}                                 & 75.50                                & 73.08                                & 68.91                                & 65.00                                 \\
SIAVC                   & 2025                  & \checkmark    &               & 73.33                                & 69.83                                 & 62.50                                 & \multicolumn{1}{c|}{56.50}                                 & 78.00                                & 75.58                                & 70.25                                & 61.83                                 \\ \hline
FixMatch                & 2020                  & \checkmark    & \checkmark    & 70.50 \textcolor{darkred}{(-1.17)}   & 66.83 \textcolor{darkgreen}{(+15.16)} & 58.17 \textcolor{darkgreen}{(+11.50)} & \multicolumn{1}{c|}{57.00 \textcolor{darkgreen}{(+12.33)}} & 74.75 \textcolor{darkgreen}{(+0.92)} & 71.50 \textcolor{darkred}{(-0.25)}   & 64.33 \textcolor{darkgreen}{(+3.91)} & 69.50 \textcolor{darkgreen}{(+19.75)}  \\
FlexMatch               & 2021                  & \checkmark    & \checkmark    & 72.00 \textcolor{darkgreen}{(+1.00)} & 70.83 \textcolor{darkgreen}{(+3.50)}  & 66.67 \textcolor{darkgreen}{(+14.34)} & \multicolumn{1}{c|}{57.50 \textcolor{darkgreen}{(+2.83)}}  & 75.50 \textcolor{darkgreen}{(+2.50)} & 74.08 \textcolor{darkred}{(-0.59)}   & 69.67 \textcolor{darkgreen}{(+6.17)} & 64.33 \textcolor{darkgreen}{(+7.92)}  \\
Dash                    & 2021                  & \checkmark    & \checkmark    & 71.67 \textcolor{darkgreen}{(+1.34)} & 71.33 \textcolor{darkgreen}{(+16.00)} & 63.83 \textcolor{darkgreen}{(+19.33)} & \multicolumn{1}{c|}{58.67 \textcolor{darkgreen}{(+18.50)}} & 72.58 \textcolor{darkred}{(-2.59)}   & 70.33\textcolor{darkred}{(-1.42)}    & 69.67\textcolor{darkgreen}{(+7.25)} & 62.75 \textcolor{darkgreen}{(+16.67)} \\
FreeMatch               & 2023                  & \checkmark    & \checkmark    & 75.50 \textcolor{darkgreen}{(+1.67)} & 72.83 \textcolor{darkgreen}{(+3.66)}  & 64.83 \textcolor{darkgreen}{(+12.50)} & \multicolumn{1}{c|}{58.67 \textcolor{darkgreen}{(+13.00)}} & 77.17 \textcolor{darkgreen}{(+2.00)} & 73.08 \textcolor{darkred}{(-0.34)}   & 68.58 \textcolor{darkred}{(-0.67)}   & 63.75 \textcolor{darkgreen}{(+2.92)}  \\
SoftMatch               & 2023                  & \checkmark    & \checkmark    & 71.17 \textcolor{darkgreen}{(-2.16)} & 69.00 \textcolor{darkred}{(-0.83)}    & 63.83 \textcolor{darkgreen}{(+1.16)}  & \multicolumn{1}{c|}{51.33 \textcolor{darkred}{(-3.00)}}    & 75.92 \textcolor{darkgreen}{(+2.92)} & 72.83 \textcolor{darkgreen}{(+1.33)} & 68.75 \textcolor{darkgreen}{(+2.42)} & 63.58 \textcolor{darkgreen}{(+3.17)}  \\
HABIT                   & 2023                  & \checkmark    & \checkmark    & 74.33\textcolor{darkgreen}{(+ 1.16)} & 71.33 \textcolor{darkgreen}{(+1.50)}  & 61.17 \textcolor{darkgreen}{(+0.17)}  & \multicolumn{1}{c|}{52.33 \textcolor{darkred}{(-2.50)}}    & 76.75 \textcolor{darkgreen}{(+1.25)} & 72.67 \textcolor{darkred}{(-0.41)}   & 69.25 \textcolor{darkgreen}{(+0.34)} & 64.83 \textcolor{darkred}{(-0.17)}    \\
SIAVC                   & 2025                  & \checkmark    & \checkmark    & 72.17 \textcolor{darkred}{(-1.16)}   & 66.83 \textcolor{darkred}{(-3.00)}    & 64.00 \textcolor{darkgreen}{(+1.50)}  & \multicolumn{1}{c|}{54.67 \textcolor{darkred}{(-1.83)}}    & 74.42 \textcolor{darkred}{(-3.58)}   & 69.75 \textcolor{darkred}{(-5.83)}   & 69.67 \textcolor{darkred}{(-0.58)}   & 64.42 \textcolor{darkgreen}{(+2.59)}  \\ \hline
MICS                    & Ours                  & \checkmark    & \checkmark    & \textbf{76.67}                       & \textbf{75.33}                        & \textbf{67.17}                        & \multicolumn{1}{c|}{\textbf{60.88}}                        & \textbf{78.67}                       & \textbf{77.92}                       & \textbf{74.33}                       & \textbf{70.25}                        \\ \hline
\end{tabular}}
\label{tab:my-table}
\end{table*}

\subsection{Datasets}
On the one hand, to verify the practical significance of the generated paired modality images. On the other hand, it is limited by publicly available paired medical image multimodal datasets. We use the public gastroscopy white light image datasets Kvasir and Kvasirv2. WtNGAN generates the narrow-band light image paired with each image, and all comparison methods use the same multimodal data. The dataset example is shown in \autoref{fig:dataset}. The Kvasir dataset contains three normal gastric anatomical structure types and three common gastric disease images. We unify the image resolution to 256×256 and divide the training set, validation set, and test set into a ratio of 6:2:2. The detailed composition of the dataset used is shown in \autoref{tab:dataset}.

\begin{figure*}[!t]
  \centering
  \includegraphics[width=0.85\linewidth]{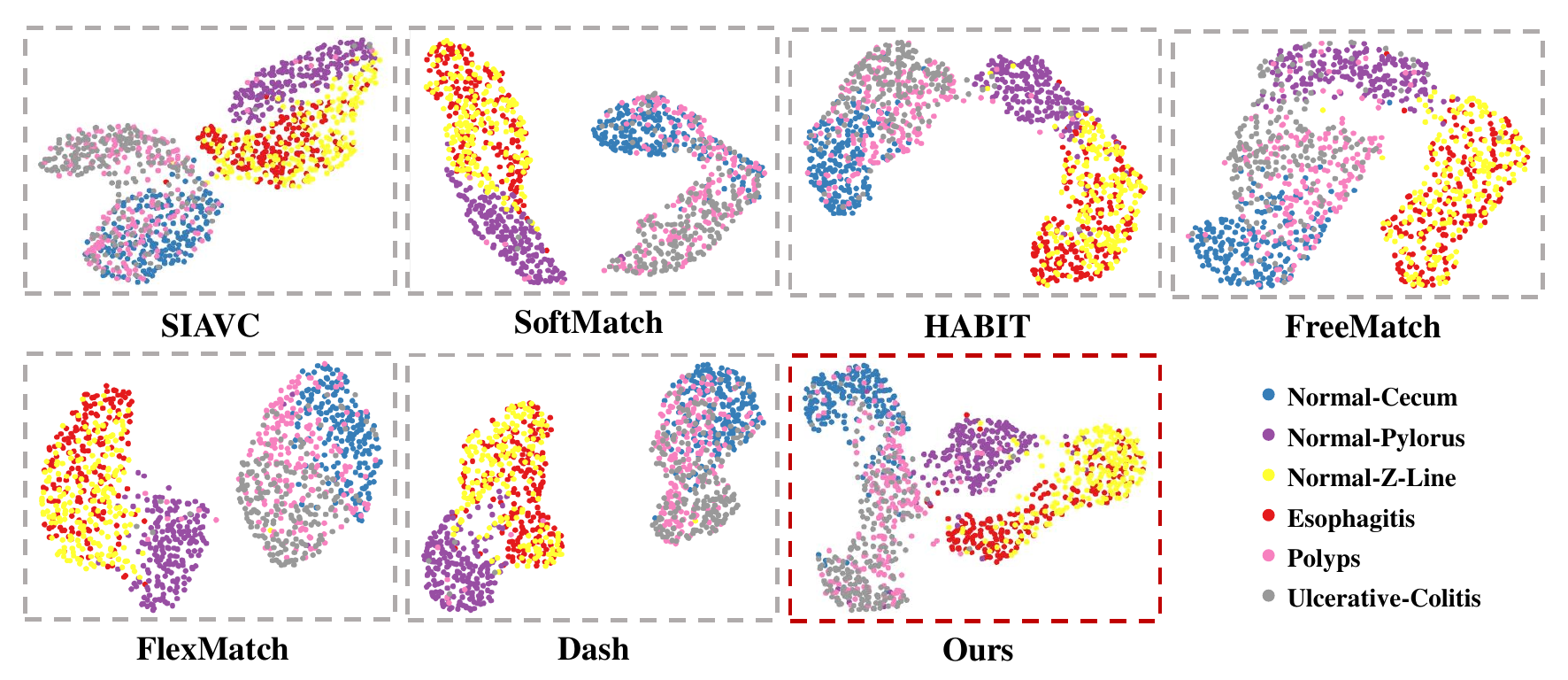}
  \caption{The UMAP (n\_components=2, n\_neighbors=30, min\_dist=1.0) feature visualization of each method on the Kvasir v2 dataset with only 5\% labeled data provided.}
\label{fig:UMAP}
\end{figure*}

\subsection{Baseline Methods}
We compare with the state-of-the-art semi-supervised classification methods trained with white light images only and with images from both modalities. The comparative methods are as follows:

\begin{itemize}

\item FixMatch \cite{fixmatch} applies consistency regularization to pseudo-labeling methods and sets a fixed confidence threshold to predict the consistency of different augmented samples.

\item FlexMatch \cite{flexmatch} dynamically adjusts the pseudo-label confidence threshold of each class of samples according to the model's learning state.

\item Dash \cite{dash} selects a subset of training samples to filter out erroneous pseudo-label samples based on the dynamic adjustment of the threshold.

%\item DebiasPL \cite{DebiasPL} eliminates classifier response and category boundary bias through counterfactual reasoning and adaptive margins.

\item FreeMatch \cite{freematch} sets a lower pseudo-label threshold in the early stage of training to accelerate model convergence and increases the threshold in the later stage of training to eliminate erroneous pseudo-label predictions.

\item SoftMatch \cite{softmatch} focuses on the quality-quantity trade-off of pseudo-label samples, maintaining numerous high-quality pseudo-labels during training to effectively utilize unlabeled samples.

\item HABIT \cite{HABIT} proposes a consistency-aware momentum heredity to alleviate bias in pseudo-label sample selection.

\item SIAVC \cite{SIAVC} proposes a super augmentation block and a cross-set augmentation module so that model optimization benefits from samples that have been well-learned.
\end{itemize}

\subsection{Quantitative and Qualitative Experimental Results}
Quantitative results are shown in \autoref{tab:my-table}. We conduct two sets of experiments under four label ratios. The results demonstrate that replacing strongly augmented images with weakly augmented ones from another modality effectively enhances semi-supervised classification performance, validating our hypothesis in Section III.B. Notably, FixMatch with multimodal training significantly improves accuracy with only 5\% labeled data, likely because it relies on abundant unlabeled samples for optimization, avoids erroneous feature representations from excessive strong augmentations, and filters out incorrect pseudo-labels with high thresholds. Other threshold-based methods show accuracy improvements ranging from 0.17\% to 18.50\%, while composite augmentation methods experience accuracy drops of 0.17\% to 5.83\% at specific label ratios. Among all results, the proposed MICS leverages multimodal data, achieving the best classification outcomes, with an accuracy of 70.25\% on the Kvasir v2 dataset using only 5\% labeled data.

The qualitative experimental results are shown in \autoref{fig:UMAP}. We provide the UMAP feature visualization results of comparative methods trained with only 5\% labeled data on the Kvasir v2 dataset. The experimental results indicate that almost all classification methods exhibit confusion between Normal-Cecum, Ulcerative, and Polyps. On the one hand, the cecum is part of the colon, and ulcers and polyps are more likely to appear in the colon region. On the other hand, with only a small amount of labeled data provided, these methods fail to focus on the lesions' characteristics and instead memorize the scenarios where the lesions appear, which is easier than the former. Notably, in the feature visualization results of the proposed method, normal cecum samples are clearly separated from ulcerative or polyps in the feature space, indicating that MICS distinguishes well between lesion locations and the lesions. MICS, SIAVC, and Dash also differentiate between the normal pyloric structure and the Z-Line. The structural similarity between the pylorus and the z-line is high, and sensitivity to the hole-like black structure may cause the model to confuse the two, resulting in low differentiation in the feature space. Finally, when labeled data is scarce, all methods fail to distinguish between Normal-Z-Line and Esophagitis. Anatomically, the Z-Line is located at the junction of the stomach and esophagus, and Esophagitis is one of the most common pathological manifestations of the Z-Line. Therefore, when Esophagitis images show low-grade inflammation or low contrast, models trained with limited labels struggle to distinguish between normal esophagus and mild esophageal inflammation.

\begin{table}[!t]
\caption{Ablation experiment of the final fine-tuned model with weights of each pre-trained component provided.}
\renewcommand{\arraystretch}{1.4}
\huge
\resizebox{0.48\textwidth}{!}{
\begin{tabular}{c|ccc|c|c}
\hline
\multicolumn{1}{c|}{\multirow{3}{*}{Train Way}} & \multicolumn{3}{c|}{Components}                                                   & Kvasir                          & Kvasirv2                        \\ \cline{2-6} 
\multicolumn{1}{c|}{}                           & \multicolumn{1}{c|}{Consistency} & \multicolumn{1}{c|}{Reconstructive} & \multicolumn{1}{c|}{Aligned}  & \multirow{2}{*}{Top-1 Acc} & \multirow{2}{*}{Top-1 Acc} \\
\multicolumn{1}{c|}{}                           & \multicolumn{1}{c|}{Learning}    & \multicolumn{1}{c|}{Learning}       & \multicolumn{1}{c|}{Learning} &                            &                                 \\ \hline
\multirow{3}{*}{Pretrain}                       & \checkmark                       &                                     &                               & 71.33\%                    & 72.50\%          \\
                                                & \checkmark                       & \checkmark                          &                               & 73.33\%                    & 75.17\%          \\
                                                & \checkmark                       & \checkmark                          & \checkmark                    & \textbf{77.04\%}           & \textbf{78.67\%}          \\ \hline
Vanilla                                         &                                  &                                     &                               & 70.33\%                    & 70.92\%          \\ \hline
\end{tabular}}
\label{tab:as_pretrain}
\end{table}

\begin{table}[!t]
\caption{Ablation studies of various fine-tuned model components using the complete pre-trained weights.}
\renewcommand{\arraystretch}{1.4}
\huge
\resizebox{0.48\textwidth}{!}{
\begin{tabular}{c|cccc|c|c}
\hline
\multirow{2}{*}{Train Way}   & \multicolumn{4}{c|}{Components}         & Kvasir                             & Kvasir v2 \\ \cline{2-7} 
                             & \multicolumn{1}{c|}{Encoder WLI}        & \multicolumn{1}{c|}{Encoder NBI}   & \multicolumn{1}{c|}{SVD}         & \multicolumn{1}{c|}{TMC}         & Top-1 Acc           & Top-1 Acc \\ \hline
\multirow{4}{*}{Fine-tuning} & \checkmark                              &                                    &                                  &                                  & 70.33\%             & 74.08\%   \\
                             & \checkmark                              & \checkmark                         &                                  &                                  & 73.33\%             & 76.91\%   \\
                             & \checkmark                              & \checkmark                         & \checkmark                       &                                  & 76.17\%             & 78.00\%   \\ \cline{2-7} 
                             & \checkmark                              & \checkmark                         & \checkmark                       & \checkmark                       & \textbf{77.04\%}    & \textbf{78.67\%}   \\ \hline
\end{tabular}}
\label{tab:as_fine}
\end{table}

\subsection{Ablation Study}
To demonstrate the contribution of each component to multimodal classification, we conduct ablation studies on the proposed method with 50\% of the labels. The ablation experiments for MICS are shown in \autoref{tab:as_pretrain} and \autoref{tab:as_fine}. \autoref{tab:as_pretrain} demonstrates the classification results of the final fine-tuned model with weights from different pre-trained components. The experimental results indicate that the synergistic learning pretraining (Pretrain) improves the classification accuracy by 6.71\% and 7.75\% on two gastroscopy datasets, respectively, compared to the model with randomly initialized weights (Vanilla). \autoref{tab:as_fine} shows the ablation results of each component during the fine-tuning phase. Compared to the single-modality WLI, introducing the NBI encoder for multimodal fusion improves the performance by 3.00\% and 2.08\% on the two gastroscopy datasets, respectively. SVD provides more generalized training samples based on the distributions of different modalities and contributes to a 2.84\% accuracy improvement on the relatively more minor Kvasir dataset. Finally, the TMC based on Dynamic Evidential fusion further enhances the classification capability of MICS by leveraging uncertainty estimation.

\section{Conclusion}
In this paper, we rethink the role of multimodal images in semi-supervised learning based on consistency regularization and propose MICS, the first semi-supervised medical image classification network to our knowledge that exploits multimodal consistency. Specifically, we use real gastroscopic WLI images and paired NBI images generated by the algorithm to classify gastroscopic diseases and anatomical structures. MICS separates the learning process of labeled data from unlabeled data, avoiding the high-confidence misprediction of unlabeled samples affecting the learning of labeled samples. MICS implements adversarial alignment pre-training of unlabeled samples and distribution shift with uncertainty fusion of a small number of labeled samples in the framework of "pre-training + fine-tuning". The proposed MICS shows promising recognition ability on gastroscopic multimodal classification datasets, especially when only small labels are provided. In the future, we will verify the superiority of MICS on more types of medical image tasks and generate more reliable multimodal data to further improve the classification accuracy.

\bibliographystyle{IEEEtran}
\bibliography{IEEEabrv,ref}

\end{document}